%% file: main.tex
\documentclass{sjtudenglab}
\usepackage{lmodern}

\usepackage{amsmath,amssymb}
\usepackage{array,tabularx,colortbl,makecell,wrapfig}
\usepackage{algorithm,algorithmic,pifont,needspace,float}

\definecolor{ourrow}{HTML}{E8F2E0}
\newcommand{\highlightrow}{\rowcolor{ourrow}}

\newcolumntype{C}{>{\centering\arraybackslash}X}
\newcommand{\cmark}{\ding{51}}
\newcommand{\xmark}{\ding{55}}
\newcommand{\algstage}[1]{\STATE \textcolor{hanblue}{\texttt{// #1}}}

\definecolor{hanblue}{rgb}{0.27,0.42,0.81}
\hypersetup{linkcolor=hanblue,citecolor=hanblue,urlcolor=hanblue,anchorcolor=hanblue}
\numberwithin{equation}{section}
\AtBeginDocument{\urlstyle{sf}}
\newcommand{\papertitle}{Draft Forcing: Target-guided Speculative Drafting and Learned Verifier for Autoregressive Video Generation}
\title{\papertitle}
\hypersetup{pdftitle={\papertitle}}
\author[1,*]{\mbox{Chenkai Xu}}
\author[1,*]{\mbox{Yuji Ren}}
\author[1]{\mbox{Zihui Liu}}
\author[1]{\mbox{Yijie Jin}}
\author[2]{\mbox{Yuhao Shen}}
\author[3]{\mbox{Xin Wang}}
\author[3]{\mbox{Yajing Chen}}
\author[3]{\mbox{Keyang Luo}}
\author[3]{\mbox{Shengju Qian}}
\author[4]{\mbox{Litong Gong}}
\author[4]{\mbox{Tiezheng Ge}}
\author[1]{\mbox{Zhijie Deng}}

\affiliation[1]{\mbox{Shanghai Jiao Tong University}}
\affiliation[2]{\mbox{Zhejiang University}}
\affiliation[3]{\mbox{LIGHTSPEED}}
\affiliation[4]{\mbox{Alibaba Group}}
\contribution[*]{\sffamily\,Equal contribution.}

\metadata[Code]{\url{https://github.com/SJTU-DENG-Lab/DraftForcing}}
\metadata[Emails]{\sffamily Chenkai Xu \email{132435xck@sjtu.edu.cn};\enspace Yuji Ren \email{renyj26@sjtu.edu.cn}}
\correspondence{\sffamily Zhijie Deng \email{zhijied@sjtu.edu.cn}}
\date{\sffamily November 26, 2026}
\abstract{\input{sec/0_abstract}}
\AtBeginDocument{
  \fancyhead[L]{\setlength{\fboxsep}{0pt}\fcolorbox{white}{white}{\includegraphics[width=3.5cm]{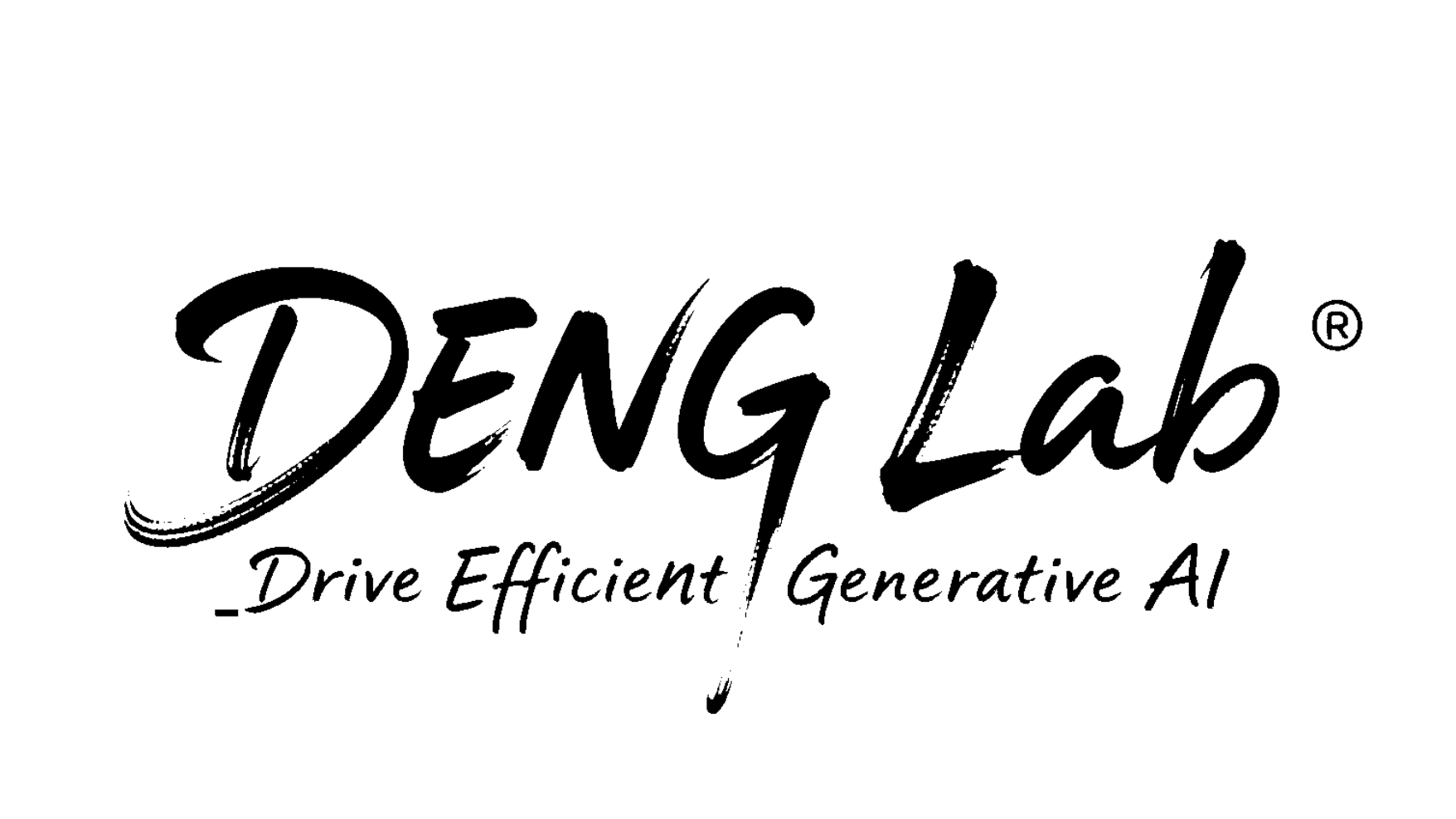}}}
  \fancyheadoffset[R]{-0.3cm}
  \fancyhead[R]{\raisebox{9pt}{\includegraphics[height=35pt]{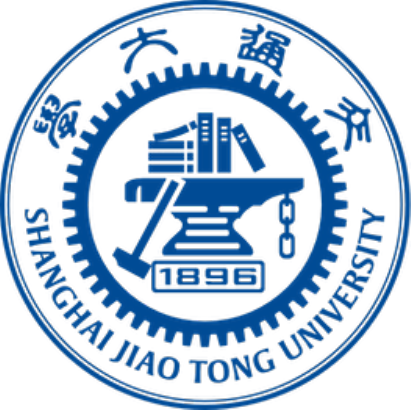}}}
}

\begin{document}
\maketitle
\begin{figure}[t]
\centering
\includegraphics[width=\linewidth]{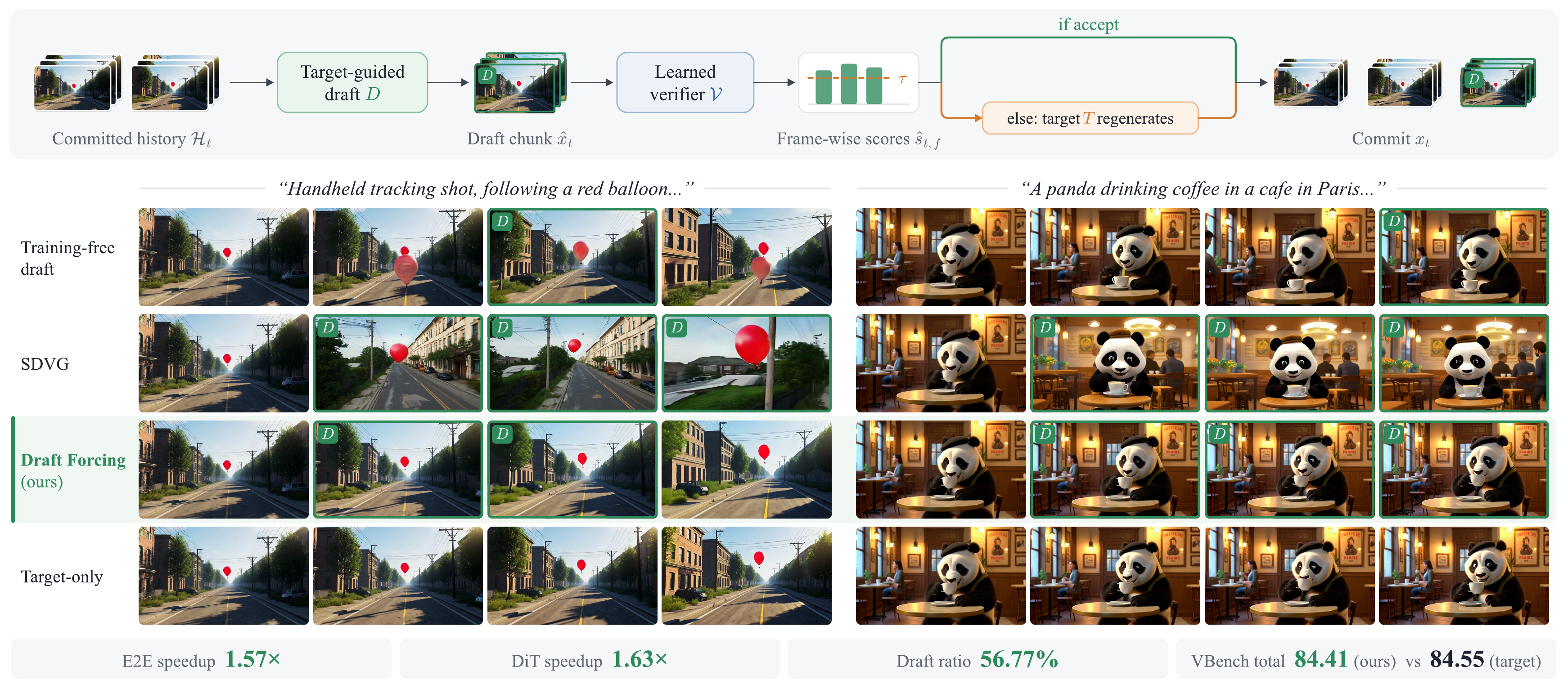}
\caption{\textbf{Draft Forcing} overview and qualitative comparison on accelerating the 14B Self-Forcing target. \textbf{Top:} chunk-wise speculation couples a target-guided draft $D$ with an on-policy trained feature-space verifier $\mathcal{V}$, accepting proposals with $\hat{s}_t \ge \tau$ or falling back to target regeneration before committing $x_t$. \textbf{Bottom:} visual comparison against Training-free Draft (accepting one off-the-shelf 1.3B draft chunk after each target chunk) and Target-only, where green ``$D$'' badges indicate draft-generated frames.}
\label{fig:qualitative}
\end{figure}

\input{sec/1_intro}
\input{sec/3_method}
\input{sec/4_experiments}
\input{sec/2_related}
\input{sec/5_conclusion}

\FloatBarrier
\setlength{\bibsep}{4pt}
\apptocmd{\thebibliography}{\interlinepenalty=10000\relax}{}{}

\beginappendix
\input{sec/appendix}
\end{document}

%% file: sec/0_abstract.tex
Autoregressive video diffusion models enable streaming synthesis chunk by chunk, yet invoking the large model for every chunk remains costly. While chunk-level speculative decoding can offload chunks to a lightweight draft model, existing approaches suffer from two limitations: (1) an off-the-shelf draft diverges from the target's trajectory, compounding errors in history; and (2) they either omit verification or rely on costly post-decode aesthetic scores rather than verifying consistency against the target. We address both with \textbf{Draft Forcing}, a framework coupling target-guided drafting with an on-policy trained verifier. Specifically, to align the draft for high acceptance, we opt for feature-level drafting, conditioning it on multi-layer target hidden states persisted from committed chunks, whose outputs are pushed to match target trajectories with multi-aspect losses. For verification, we propose an on-policy trained feature-space verifier predicting CDF-calibrated frame-wise relative RMSE against the target. It is trained on speculative rollouts as contexts anneal from random to inference-aligned mixtures, and gates proposals without VAE decoding. On the 14B Self-Forcing target, Draft Forcing achieves $1.63\times$ DiT and $1.57\times$ end-to-end speedups on VBench with minimal quality loss, and up to $1.93\times$ DiT ($1.82\times$ end-to-end) when combined with orthogonal KV compression. It generalizes to accelerate a streaming variant of MiniMax-H3, generating over $50\%$ draft frames with competitive quality.

%% file: sec/1_intro.tex
\section{Introduction}
\label{sec:intro}

Autoregressive video diffusion models have emerged as a promising paradigm for streaming video generation, as they synthesize video chunk by chunk in causal order~\cite{chen2024diffusionforcing,yin2025causvid,huang2025selfforcing}. By reusing historical KV caches and performing few-step denoising per chunk, recent streaming models achieve high single-chunk efficiency~\cite{yin2025causvid,huang2025selfforcing,krea2025realtime}. However, inference still requires invoking the large model sequentially for every chunk, expending the same heavy compute on smooth, predictable continuations as on complex motion transitions.

To reduce this compute overhead, speculative decoding~\cite{leviathan2023speculative,chen2023speculative,cai2024medusa,li2024eagle,li2025eagle3,chen2026dflash,dflare2026} offers a natural remedy, where a lightweight draft model proposes candidates to be either accepted or regenerated by the large target model. Existing video speculative methods fall into two categories: timestep-level approaches~\cite{dts2026,flowcast2026,asdsv2025} speculate across denoising steps, yielding limited gains in few-step streaming models; and chunk-level approaches~\cite{nextforcing2026,sdvg2026} speculate across chunks. However, chunk-level speculation faces two key challenges: (1) an off-the-shelf draft~\cite{sdvg2026} diverges from the target's trajectory, causing errors to compound once committed into the history; and (2) continuous video chunks cannot be verified in parallel like discrete tokens in language models, while existing methods either omit verification~\cite{nextforcing2026} or rely on costly post-decode aesthetic scores~\cite{sdvg2026}, which cannot measure consistency with the target.

\begin{figure}[t]
\centering
\includegraphics[width=\textwidth]{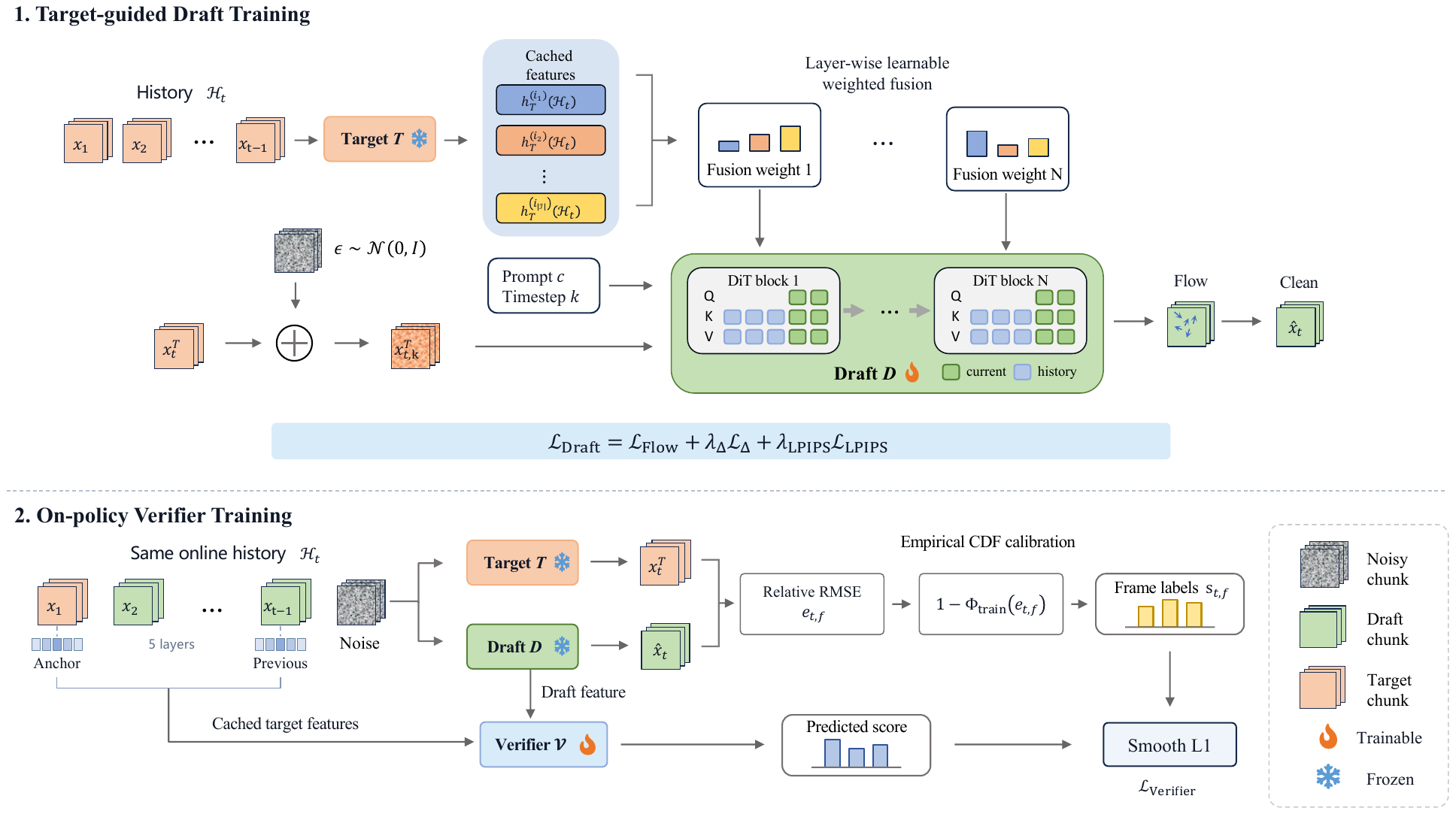}
\caption{Training of \textbf{Draft Forcing}. (1) Target-guided draft training fuses cached target hidden states persisted from committed historical chunks and combines flow matching, temporal difference, and LPIPS losses. (2) On-policy verifier training pairs draft and target outputs along speculative rollouts to learn frame-wise CDF-normalized discrepancy scores.}
\label{fig:training}
\end{figure}

To address both challenges, we present \textbf{Draft Forcing}, a chunk-level speculative framework that couples target-guided drafting with an on-policy trained verifier (Figures~\ref{fig:training} and~\ref{fig:inference}). To align the draft with the target for high acceptance, we follow the feature-level drafting paradigm~\cite{li2024eagle,li2025eagle3,chen2026dflash,dflare2026}. Using layer-wise fusion and dedicated context KV projections, each draft layer conditions on multi-layer target hidden states persisted from committed chunks. On collected target trajectories, the draft outputs are pushed to match the target from multiple aspects. Beyond the standard flow-matching loss~\cite{lipman2023flowmatching}, we introduce a temporal latent difference loss $\mathcal{L}_{\Delta}$ to encourage accurate fitting of dynamic regions and temporal motion, alongside a decoded-crop perceptual LPIPS loss~\cite{zhang2018lpips} to sharpen spatial details.

For low-cost and effective verification, we propose an on-policy trained feature-space verifier, which predicts frame-wise scores to determine chunk acceptance. To ensure that the verifier's training contexts match the mixed draft--target histories at inference, we train the verifier on-policy during its own speculative rollouts, progressively annealing the training contexts from random mixtures to inference-aligned histories. Specifically, the verifier takes cached target history features and draft proposal features as input, and predicts the relative RMSE between the draft proposal and the target reference as a measure of consistency. This relative RMSE is normalized by its empirical CDF into a score for controllable draft acceptance ratios. At inference, the verifier gates proposals directly in feature space without VAE decoding, accepting chunks whose worst-frame score exceeds a threshold $\tau$ and regenerating rejected chunks with the target.

To avoid costly training from scratch, we initialize the draft's DiT~\cite{peebles2023dit} backbone from the pretrained 1.3B Self-Forcing model~\cite{huang2025selfforcing} and fine-tune only lightweight LoRA~\cite{hu2021lora} adapters and layer-wise fusion modules. When accelerating the 14B Self-Forcing target~\cite{huang2025selfforcing,krea2025realtime} with $22.5\text{k}$ draft training steps and $700$ on-policy verifier training steps, Draft Forcing achieves $1.63\times$ DiT and $1.57\times$ end-to-end speedups on VBench~\cite{huang2024vbench} with minimal quality loss, and up to $2.02\times$ DiT ($1.89\times$ end-to-end) under a more aggressive setting. Moreover, without additional training, Draft Forcing is naturally orthogonal to KV-cache compression methods~\cite{ji2026forcingkv,guo2026dummyforcing}, reaching up to $1.93\times$ DiT and $1.82\times$ end-to-end speedups while preserving visual quality when integrated with Dummy Forcing~\cite{guo2026dummyforcing}. Finally, Draft Forcing generalizes across model families via a learned latent mapper~\cite{li2026solh3}. When applied to a streaming variant~\cite{lu2026raven} of the powerful open-source video diffusion model MiniMax-H3~\cite{minimax2026h3}, with $8.5\text{k}$ draft training steps and $540$ on-policy verifier training steps, it generates over $50\%$ of frames with the draft while maintaining competitive generation quality.

In summary, our main contributions are:
\begin{itemize}
  \item \textbf{Target-guided Drafting:} Following the feature-level drafting paradigm, we condition the draft model on multi-layer target hidden states persisted from committed historical chunks via dedicated context KV projections, and train it on target trajectories with flow-matching, temporal difference, and perceptual LPIPS losses.
  \item \textbf{On-policy Trained Verifier:} We propose an on-policy trained feature-space verifier supervised along its own speculative rollouts to predict CDF-normalized relative RMSE scores against the target, which enables low-cost verification in feature space without VAE decoding.
  \item \textbf{Effectiveness and Cross-backbone Generalization:} Reusing a pretrained 1.3B Self-Forcing draft with LoRA tuning on the 14B Self-Forcing target, Draft Forcing achieves speedups of $1.63\times$ DiT and $1.57\times$ end-to-end alone, and up to $1.93\times$ DiT and $1.82\times$ end-to-end when combined with orthogonal KV compression~\cite{ji2026forcingkv,guo2026dummyforcing}, with minimal quality loss on VBench. Moreover, it generalizes across model families to accelerate a streaming variant~\cite{lu2026raven} of MiniMax-H3~\cite{minimax2026h3} with over $50\%$ draft frames.
\end{itemize}

%% file: sec/3_method.tex
\section{Preliminaries}
\label{sec:prelim}

\subsection{Streaming Autoregressive Video Generation}
\label{sec:prelim-streaming}

Autoregressive video diffusion models enable streaming video synthesis by generating chunks sequentially in causal order~\cite{chen2024diffusionforcing,yin2025causvid,huang2025selfforcing}. Operating in the VAE latent space~\cite{rombach2022ldm,wan2025}, a streaming model partitions a video into $N$ chunks $x_1,\ldots,x_N$ of $F$ frames each. Conditioned on a text prompt $c$ and the committed history $\mathcal{H}_t=\{x_1,\ldots,x_{t-1}\}$ ($\mathcal{H}_1=\emptyset$), it factorizes the joint distribution as
\begin{equation}
p(x_1,\ldots,x_N\mid c)=\prod_{t=1}^{N}p(x_t\mid\mathcal{H}_t,c).
\label{eq:causal-chunk}
\end{equation}
Reusing cached KV states of $\mathcal{H}_t$ with few-step denoising~\cite{yin2025causvid,huang2025selfforcing,causalforcing2026,causalrcm2026} achieves high single-chunk efficiency. Yet Eq.~\eqref{eq:causal-chunk} still invokes the large model sequentially for every chunk, expending uniform compute on both smooth continuations and complex motion transitions. This redundancy motivates chunk-level speculative decoding.

\subsection{Speculative Decoding from Language to Video}
\label{sec:prelim-llm}

\begin{wrapfigure}{r}{0.46\textwidth}
\centering
\includegraphics[width=\linewidth]{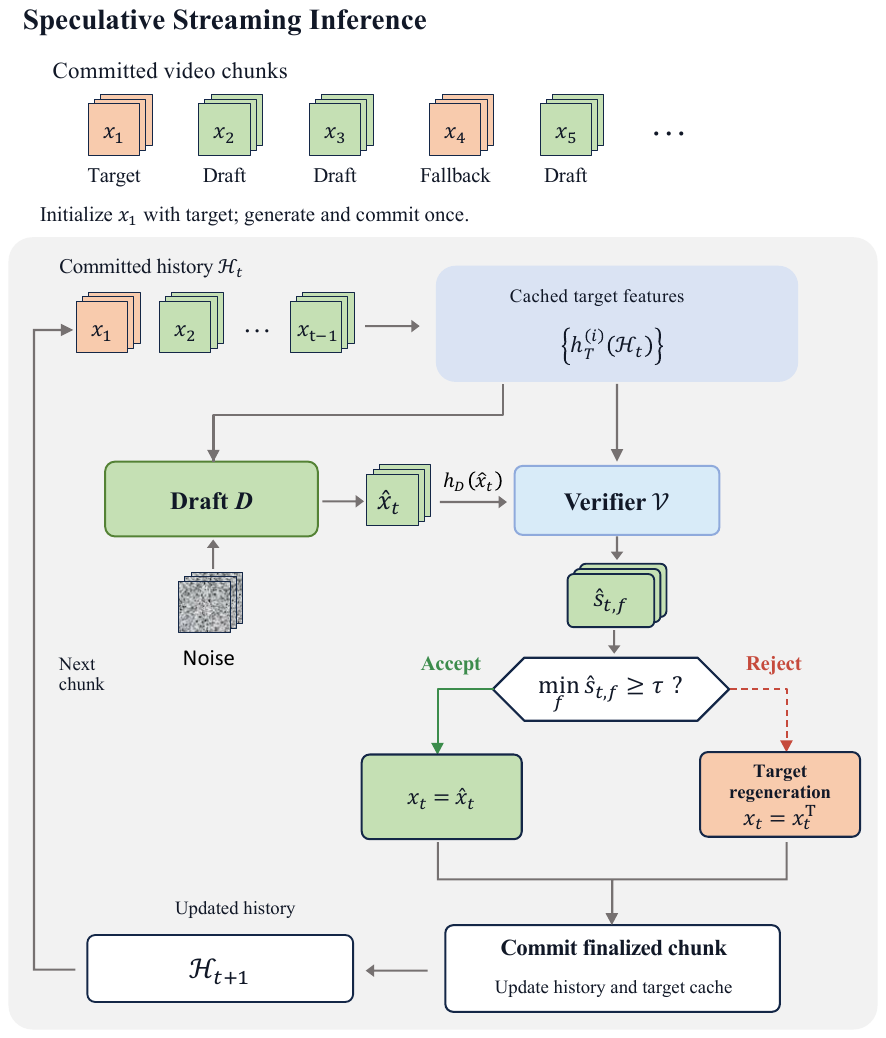}
\caption{Speculative streaming inference of \textbf{Draft Forcing}. The verifier accepts a draft proposal or invokes target regeneration, and the finalized chunk updates the committed history and target cache. Appendix~\ref{app:inference} specifies cumulative-error-budget routing.}
\label{fig:inference}
\end{wrapfigure}
In autoregressive language modeling, speculative decoding~\cite{leviathan2023speculative,chen2023speculative} relies on low-cost parallel verification. A lightweight draft model proposes discrete candidate tokens, which the large target model $T$ verifies in one parallel forward pass via rejection sampling. Since per-token compute is small, parallel verification yields substantial speedups under low concurrency.

Extending this paradigm to streaming video, however, faces distinct challenges. Timestep-level methods~\cite{asdsv2025,dts2026,flowcast2026} speculate across denoising steps within a chunk, offering limited gains in few-step streaming models ($1$--$4$ steps). Chunk-level methods instead speculate across chunks, yet face two key challenges: (1) an off-the-shelf lightweight draft~\cite{sdvg2026} diverges from $T$'s trajectory, compounding errors once committed into $\mathcal{H}_t$; and (2) continuous chunks involve heavy spatiotemporal computation, so invoking $T$ to verify a drafted chunk erases the speedup. Consequently, prior chunk-level methods either omit verification~\cite{nextforcing2026}, risking error accumulation, or verify decoded frames via ImageReward~\cite{sdvg2026,xu2023imagereward}, incurring costly VAE decoding while measuring aesthetics rather than consistency with $T$.

\section{Method}
\label{sec:method}

To address both challenges, we present \textbf{Draft Forcing}, a chunk-level speculative framework that couples target-guided drafting with an on-policy trained verifier. Below, we detail target-guided drafting (Section~\ref{sec:draft}), the on-policy trained verifier (Section~\ref{sec:scorer}), speculative inference (Section~\ref{sec:commit}), and cross-backbone adaptation (Section~\ref{sec:cross-model}).

\subsection{Target-guided Drafting}
\label{sec:draft}

To align the lightweight draft $D$ with the target $T$ for high acceptance (Figure~\ref{fig:training}), we follow the paradigm of feature-level drafting~\cite{li2024eagle,li2025eagle3,chen2026dflash,dflare2026}, conditioning $D$ on multi-layer target hidden states persisted from committed historical chunks. When denoising chunk $t$, each draft layer $\ell$ uses its own set of learnable weights to compute a weighted sum of the cached hidden states from selected target layers:
\begin{equation}
f^{(\ell)}_{\mathrm{ctx}}=\mathrm{Fuse}_\ell\bigl(\{h^{(i)}_T(\mathcal{H}_t)\}_{i\in\mathcal{I}}\bigr),
\label{eq:fusion}
\end{equation}
where $h^{(i)}_T(\mathcal{H}_t)$ denotes the cached hidden states of the committed history $\mathcal{H}_t$ at target layer $i$, and $\mathcal{I}$ is a selected subset of target layers. Because $f^{(\ell)}_{\mathrm{ctx}}$ encodes clean committed history from $T$ whereas draft hidden states $z^{(\ell)}$ represent noisy latents along the current chunk's denoising trajectory, sharing KV projections would conflate two distinct distributions. Motivated by~\cite{dflare2026}, we maintain dedicated context projections $(W^{(\ell)}_{K,\mathrm{ctx}},\,W^{(\ell)}_{V,\mathrm{ctx}})$ for $f^{(\ell)}_{\mathrm{ctx}}$, separate from the current-chunk self-attention projections $(W^{(\ell)}_{Q,\mathrm{cur}},\,W^{(\ell)}_{K,\mathrm{cur}},\,W^{(\ell)}_{V,\mathrm{cur}})$:
\begin{equation}
\begin{aligned}
K^{(\ell)} &= \bigl[f^{(\ell)}_{\mathrm{ctx}}W^{(\ell)}_{K,\mathrm{ctx}};\;z^{(\ell)}W^{(\ell)}_{K,\mathrm{cur}}\bigr], \\
V^{(\ell)} &= \bigl[f^{(\ell)}_{\mathrm{ctx}}W^{(\ell)}_{V,\mathrm{ctx}};\;z^{(\ell)}W^{(\ell)}_{V,\mathrm{cur}}\bigr],
\end{aligned}
\label{eq:het-kv}
\end{equation}
with causal queries $Q^{(\ell)}=z^{(\ell)}W^{(\ell)}_{Q,\mathrm{cur}}$.

To push $D$'s outputs to match the target trajectories from multiple aspects, we first collect clean trajectories $X^T=\{x^T_1,\ldots,x^T_N\}$ generated by the frozen target $T$. For the $t$-th target chunk $x^T_t$, we sample a denoising step $k$ from the few-step schedule~\cite{huang2025selfforcing}, and perturb $x^T_t$ with Gaussian noise $\epsilon\sim\mathcal{N}(0,I)$ via forward flow matching to obtain the noisy chunk $x^T_{t,k}$ (Appendix~\ref{app:flow-schedule}). Conditioned on $c$ and the fused context $f_{\mathrm{ctx}}=\{f^{(\ell)}_{\mathrm{ctx}}\}_{\ell}$, $D$ is trained with the flow-matching loss~\cite{lipman2023flowmatching,liu2023rectifiedflow}:
\begin{equation}
\mathcal{L}_{\mathrm{Flow}}=\mathbb{E}_{k,x^T_t,\epsilon}\Bigl[\bigl\|v_D(x^T_{t,k},k,c,f_{\mathrm{ctx}})-(\epsilon-x^T_t)\bigr\|_2^2\Bigr],
\label{eq:flow-loss}
\end{equation}
where $v_D(\cdot)$ is the velocity predicted by $D$. To encourage accurate fitting of dynamic regions and temporal motion, we introduce a temporal difference loss between $D$'s denoised chunk $\hat{x}_t$ and $x^T_t$:
\begin{equation}
\mathcal{L}_{\Delta}=\frac{1}{F}\sum_{f=1}^{F}\mathbb{E}_{k,x^T_t,\epsilon}\bigl[d_{\mathrm{Charb}}\bigl(\Delta\hat{x}_{t,f},\,\Delta x^T_{t,f}\bigr)\bigr],
\label{eq:delta-loss}
\end{equation}
where $f\in\{1,\ldots,F\}$ indexes the frames within chunk $t$, $\Delta\hat{x}_{t,f}=\hat{x}_{t,f}-\hat{x}_{t,f-1}$ and $\Delta x^T_{t,f}=x^T_{t,f}-x^T_{t,f-1}$ denote adjacent-frame latent differences (with frame $0$ denoting the last frame of $x^T_{t-1}$), and $d_{\mathrm{Charb}}(\cdot,\cdot)$ denotes the Charbonnier distance~\cite{charbonnier1997deterministic}. To further sharpen spatial details, following~\cite{wan2025,wu2026pssr}, we decode random spatial crops of $\hat{x}_t$ and $x^T_t$ through the frozen VAE to compute a frame-wise LPIPS loss $\mathcal{L}_{\mathrm{LPIPS}}$~\cite{zhang2018lpips}, yielding the overall draft objective:
\begin{equation}
\mathcal{L}_{\mathrm{Draft}}=\mathcal{L}_{\mathrm{Flow}}+\lambda_{\Delta}\,\mathcal{L}_{\Delta}+\lambda_{\mathrm{LPIPS}}\,\mathcal{L}_{\mathrm{LPIPS}},
\label{eq:draft-loss}
\end{equation}
where $\lambda_{\Delta}$ and $\lambda_{\mathrm{LPIPS}}$ are trade-off weights.

\subsection{On-policy Trained Verifier}
\label{sec:scorer}

For low-cost and effective verification, we propose an on-policy trained feature-space verifier $\mathcal{V}$ that predicts frame-wise consistency scores $\{\hat{s}_{t,f}\}_{f=1}^{F}\in[0,1]^F$ between the draft proposal $\hat{x}_t$ and the target $T$. To capture both local transition continuity and multi-chunk drift without attending over the entire history, $\mathcal{V}$ extracts cached target features $\{h^{(i)}_T(\mathcal{H}_t^{\mathcal{V}})\}_{i\in\mathcal{I}}$ from two key historical chunks $\mathcal{H}_t^{\mathcal{V}}=\{x_{t-1},x_{t_{\mathrm{anc}}}\}$, where $x_{t-1}$ is the preceding committed chunk and $t_{\mathrm{anc}}<t$ is the index of the latest chunk generated by $T$. Conditioned on these historical features and the draft's final-layer hidden state $h_D(\hat{x}_t)$, $\mathcal{V}$ predicts $\{\hat{s}_{t,f}\}_{f=1}^{F}$ in a single forward pass.

To ensure that the contexts received by $\mathcal{V}$ during training match the mixed draft--target distribution at inference~\cite{bengio2015scheduled,ross2011dagger}, we train $\mathcal{V}$ on-policy during its own speculative rollouts (Figure~\ref{fig:training}), where $\mathcal{H}_t$ advances autoregressively according to $\mathcal{V}$'s acceptance decisions. As $\mathcal{V}$ improves, the training rollouts progressively anneal from random mixtures to inference-aligned contexts. At each chunk $t$ along the rollout, we generate both the draft proposal $\hat{x}_t$ and a paired target reference $x^T_t$, and measure their frame-wise discrepancy via the relative RMSE:
\begin{equation}
e_{t,f}=\sqrt{\frac{\mathrm{mean}\bigl((\hat{x}_{t,f}-x^T_{t,f})^2\bigr)}{\mathrm{mean}\bigl((x^T_{t,f})^2\bigr)+\varepsilon}},
\label{eq:relative-rmse}
\end{equation}
where $\mathrm{mean}(\cdot)$ averages over the channel and spatial dimensions of frame $f$, and $\varepsilon$ is a small constant for numerical stability. To allow a single threshold $\tau$ to control acceptance across prompts and chunk positions, we calibrate $e_{t,f}$ into $[0,1]$ via its empirical CDF $\Phi_{\mathrm{train}}$ tracked over on-policy rollouts, yielding the target score:
\begin{equation}
s_{t,f}=1-\Phi_{\mathrm{train}}(e_{t,f}),
\label{eq:cdf-score}
\end{equation}
where larger $s_{t,f}$ indicates higher consistency with $T$. The verifier $\mathcal{V}$ is then trained by minimizing the following loss:
\begin{equation}
\mathcal{L}_{\mathrm{Verifier}}=\frac{1}{F}\sum_{f=1}^{F}d_{\mathrm{Smooth}L_1}\bigl(\hat{s}_{t,f},\,s_{t,f}\bigr),
\label{eq:scorer-loss}
\end{equation}
where $d_{\mathrm{Smooth}L_1}(\cdot,\cdot)$ denotes the Smooth $L_1$ distance~\cite{girshick2015fastrcnn}. In contrast, if $\mathcal{V}$ were supervised off-policy on pre-recorded pure-target trajectories $X^T$, any earlier accepted draft chunk would shift the subsequent history $\mathcal{H}_t$ away from the offline trajectory, biasing $\mathcal{V}$ toward assigning lower scores to later chunks and failing to reflect actual chunk quality. Appendix~\ref{app:rmse-analysis} analyzes the verifier's fitting quality.

\subsection{Combined Speculative Inference}
\label{sec:commit}

As summarized in Figure~\ref{fig:inference} and Algorithm~\ref{alg:draftforcing} (Appendix~\ref{app:inference}), Draft Forcing accelerates streaming inference by dynamically alternating between lightweight draft proposals and target generation. After the target generates the initial chunk $x_1$, at each subsequent chunk $t$, $D$ proposes a candidate $\hat{x}_t$ and $\mathcal{V}$ predicts its frame-wise scores $\{\hat{s}_{t,f}\}_{f=1}^{F}$ in feature space. Because artifacts on a single frame can disrupt temporal continuity, we define the chunk-level acceptance score via worst-frame selection~\cite{sdvg2026}:
\begin{equation}
\hat{s}_t=\min_{f\in\{1,\ldots,F\}}\hat{s}_{t,f}.
\label{eq:chunk-score}
\end{equation}

Given an acceptance threshold $\tau\in[0,1]$, proposals with $\hat{s}_t\ge\tau$ are accepted in place of $T$, whereas rejected chunks are regenerated by $T$, after which the finalized chunk updates the target cache for subsequent chunks.

Even when individual proposals pass the threshold $\tau$, accepting many draft chunks consecutively can gradually accumulate approximation errors. To bound multi-chunk drift, we introduce a budget-triggered target refresh mechanism that invokes $T$ once the accumulated relative RMSE across consecutive accepted draft chunks reaches a budget $\mathcal{B}$, where each frame's relative RMSE is estimated from its predicted score via $\Phi_{\mathrm{train}}^{-1}(1-\hat{s}_{t,f})$.

\subsection{Cross-backbone Speculative Adaptation}
\label{sec:cross-model}

When $D$ and $T$ come from different model families (e.g., accelerating RAVEN-H3~\cite{lu2026raven,minimax2026h3} with a Wan-based draft), their VAE latent spaces differ, so $D$'s proposal $\hat{x}^D_t$ cannot be directly committed into $T$. Inspired by decoder-free latent-to-latent mapping~\cite{li2026solh3}, we adapt this design across heterogeneous VAEs by encoding the same videos with $D$'s and $T$'s VAEs to obtain paired latents $(x^D,x^T)$. We first pretrain a lightweight latent mapper $\mathcal{M}$ in latent space and then fine-tune it with a decoder-aware objective (see Appendix~\ref{app:training} for details):
\begin{equation}
\begin{aligned}
\mathcal{L}_{\mathrm{mapper}} &= d_{L_2}\bigl(\mathcal{M}(x^D),\,x^T\bigr) \\
&\quad + \lambda_{\mathrm{dec}}\,d_{L_2}\bigl(\mathcal{D}_T(\mathcal{M}(x^D)),\,\mathcal{D}_T(x^T)\bigr),
\end{aligned}
\label{eq:mapper-loss}
\end{equation}
where $\mathcal{D}_T$ is $T$'s frozen VAE decoder, $d_{L_2}(\cdot,\cdot)$ denotes the $L_2$ distance, and $\lambda_{\mathrm{dec}}$ is a trade-off weight. At inference, each draft proposal $\hat{x}^D_t$ is mapped into $T$'s latent space via $\hat{x}_t=\mathcal{M}(\hat{x}^D_t)$ before commitment.

%% file: sec/4_experiments.tex
\section{Experiments}
\label{sec:experiments}

This section details the experimental setup and evaluates the generation quality, acceleration efficacy, and cross-backbone generalizability of Draft Forcing.

\subsection{Experimental Setup}
\label{sec:setup}

\noindent\textbf{Models.} Our primary setting pairs the frozen 14B Self-Forcing target (Krea Realtime 14B~\cite{krea2025realtime}) with a pretrained 1.3B Self-Forcing draft~\cite{huang2025selfforcing}, using five-step and four-step sampling, respectively. Each video comprises seven chunks of three latent frames, decoded into $81$ RGB frames at $480\times832$ resolution and $16$ fps. We adapt the draft with LoRA and target-feature fusion, then train the verifier on speculative rollouts with both generative models frozen. We additionally evaluate cross-backbone adaptation to the four-step RAVEN-H3~\cite{lu2026raven,minimax2026h3}, the RAVEN streaming adaptation~\cite{lu2026raven} of MiniMax-H3~\cite{minimax2026h3}, through feature alignment and latent mapping. Training configurations and architecture details are provided in Appendix~\ref{app:training}.

\noindent\textbf{Benchmarks and Metrics.} For the primary setting, we evaluate on VBench~\cite{huang2024vbench}, reporting total, quality, semantic, and dynamic degree (DD) scores in Table~\ref{tab:vbench-full} and all $16$ dimensions in Appendix~\ref{app:full-vbench}. We compare target-only, Draft Forcing, two speculative baselines---training-free draft and SDVG~\cite{sdvg2026}---and two recent KV-cache compression methods, Forcing-KV~\cite{ji2026forcingkv} and Dummy Forcing~\cite{guo2026dummyforcing}, both alone and combined with Draft Forcing. Cross-backbone experiments and ablation studies use the $128$-prompt MovieGenVideoBench subset~\cite{polyak2024moviegen} under the CausVid protocol~\cite{yin2025causvid}, reporting temporal quality, frame quality, and text alignment. Draft ratio denotes the fraction of draft-generated chunks on Self-Forcing and output frames on RAVEN-H3. The primary, cross-backbone, and ablation settings report SSIM~\cite{wang2004ssim}, PSNR, and LPIPS~\cite{zhang2018lpips} against target-only outputs generated with the same prompt and initial noise. Across all three experimental settings, Draft Forcing uses $\tau{=}0.25$ and $\mathcal{B}{=}1.6$ by default, with a target refresh on the next chunk once the cumulative error budget is reached. On the uncompressed target, we additionally evaluate a more aggressive setting, keeping all other inference settings fixed.

\noindent\textbf{Efficiency.} Videos are generated at batch size $1$. We measure DiT sampling and end-to-end (E2E) speedups relative to target-only generation using paired prompts and seeds with synchronous chunk-wise VAE decoding. DiT time covers drafting, verification, fallback target generation, and KV operations. E2E time spans text encoding, DiT sampling, VAE decoding, and RGB transfer to CPU. Detailed measurement boundaries are given in Appendix~\ref{app:measurement}.

\subsection{Main Results}
\label{sec:main-results}

\input{sec/tables_lightx2v_main}

\noindent\textbf{Standalone Acceleration.} As shown in Table~\ref{tab:vbench-full} and Figure~\ref{fig:qualitative}, Draft Forcing accelerates the 14B target while preserving near-target VBench quality and visual fidelity. Under the default setting, it achieves $1.63\times$ DiT and $1.57\times$ E2E speedups with minimal quality loss, scoring $84.41$ against $84.55$ for target-only on VBench total. It also attains the highest target consistency among all compared methods with an SSIM of $0.6764$, outperforming standalone Forcing-KV and Dummy Forcing in speed. Under the aggressive setting, raising the draft ratio to $70.79\%$ further pushes speedups to $2.02\times$ DiT and $1.89\times$ E2E while keeping VBench total at $84.35$. Per-chunk acceptance and routing analysis is provided in Appendix~\ref{app:chunk-analysis}.

\noindent\textbf{Comparison with Speculative Baselines.} Existing chunk-level speculative baselines suffer clear fidelity or quality degradation compared with Draft Forcing. Training-free draft yields only a $1.35\times$ E2E speedup and exhibits noticeable trajectory drift, as reflected by its lower SSIM. While SDVG reaches a $1.67\times$ E2E speedup, its VBench total drops to $82.76$ and its SSIM falls to $0.4660$. Beyond relying on an unaligned draft and aesthetic filtering, SDVG updates the draft KV cache with its own proposals rather than committed target chunks, leaving the draft on an independent trajectory and causing abrupt visual jumps across transitions. By contrast, our aggressive setting achieves a faster $1.89\times$ E2E speedup while preserving near-target quality and substantially higher fidelity.

\noindent\textbf{Compatibility with KV Compression.} Draft Forcing integrates orthogonally with KV-cache compression to deliver additive acceleration. When combined with Forcing-KV and Dummy Forcing, it brings an additional $1.50\times$ and $1.48\times$ E2E speedup over each KV-only baseline, reaching $1.93\times$ DiT and $1.82\times$ E2E speedups alongside Dummy Forcing. Across both combinations, VBench total remains virtually unchanged relative to target-only with only a modest drop in SSIM, confirming that reducing target calls and compressing KV caches provide complementary gains.

\begin{figure}[t]
\centering
\includegraphics[width=\textwidth]{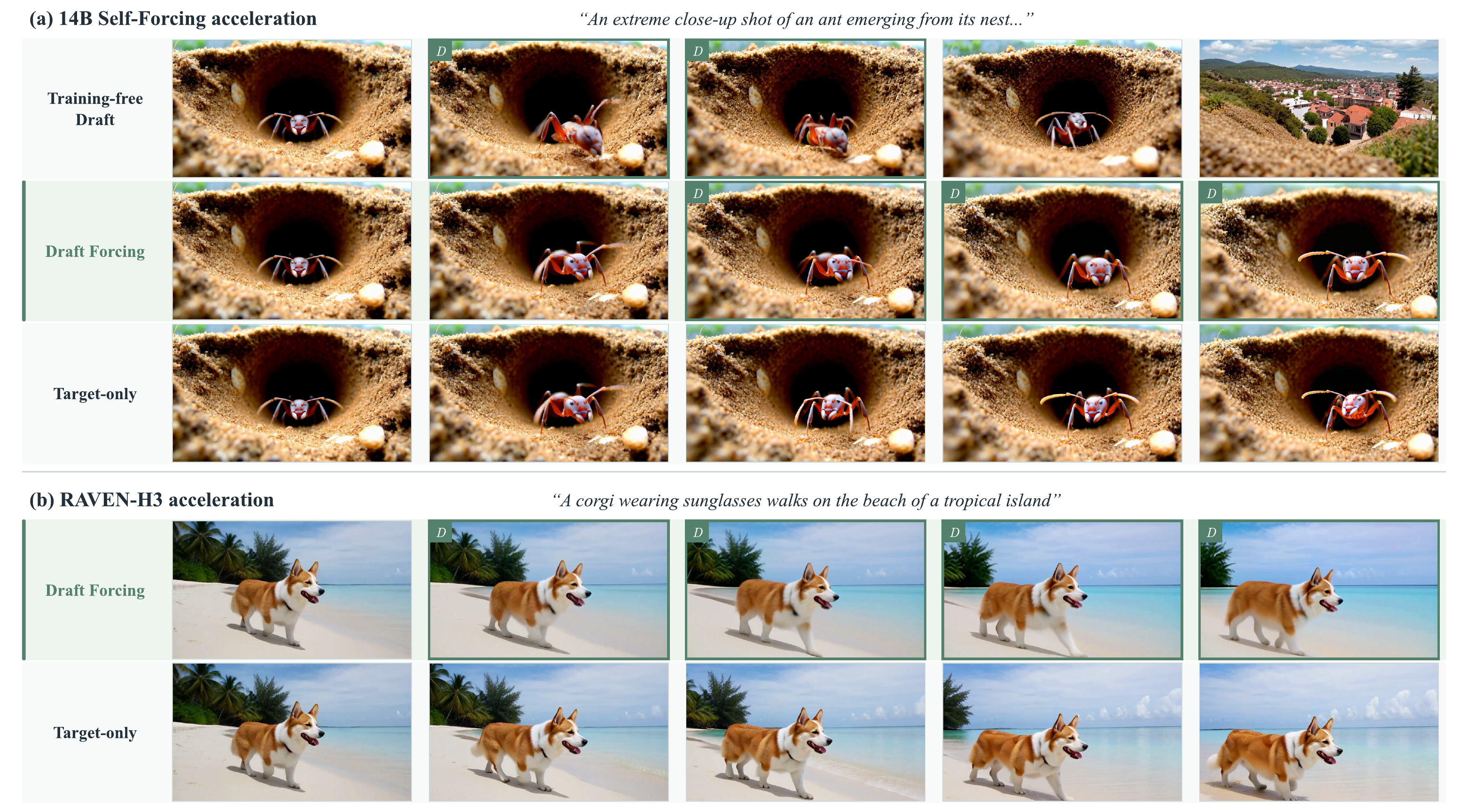}
\caption{Qualitative comparison across backbones. \textbf{Top (a):} accelerating the 14B Self-Forcing target against Training-free Draft and Target-only. \textbf{Bottom (b):} cross-backbone acceleration on the RAVEN-H3 target against Target-only. Green ``$D$'' badges indicate draft-generated frames.}
\label{fig:qualitative-summary}
\end{figure}

\begin{table}[H]
\centering
\small
\setlength{\aboverulesep}{0pt}
\setlength{\belowrulesep}{0pt}
\setlength{\extrarowheight}{2.4pt}
\setlength{\tabcolsep}{2.6pt}
\begin{tabularx}{\textwidth}{lCCCCCCC}
\toprule
Method & \begin{tabular}[t]{@{}c@{}}Temporal\\Quality$\uparrow$\end{tabular} & \begin{tabular}[t]{@{}c@{}}Frame\\Quality$\uparrow$\end{tabular} & \begin{tabular}[t]{@{}c@{}}Text\\Alignment$\uparrow$\end{tabular} & SSIM$\uparrow$ & PSNR$\uparrow$ & LPIPS$\downarrow$ & \begin{tabular}[t]{@{}c@{}}Draft\\Ratio\end{tabular} \\
\midrule
Target-only (RAVEN-H3) & 91.13 & 64.33 & 25.96 & -- & -- & -- & -- \\
\highlightrow
\textbf{Draft Forcing} & 89.74 & 60.41 & 25.96 & 0.6818 & 22.65 & 0.3355 & 51.99\% \\
\bottomrule
\end{tabularx}
\caption{Cross-backbone adaptation results on the RAVEN-H3 target using the 1.3B Self-Forcing (Wan2.1) draft, evaluated on the 128-prompt MovieGenVideoBench subset~\cite{polyak2024moviegen} under the CausVid protocol~\cite{yin2025causvid} using VBench~\cite{huang2024vbench}. SSIM, PSNR, and LPIPS are measured against target-only videos from the same prompts. Draft ratio denotes the percentage of final output frames generated by the draft model.}
\label{tab:h3}
\end{table}

\noindent\textbf{Cross-backbone Adaptation.} Table~\ref{tab:h3} evaluates the Wan-based $1.3$B draft on the heterogeneous RAVEN-H3 target via feature adaptation and latent mapping. Draft Forcing offloads $51.99\%$ of output frames to the draft while matching the target's text alignment and preserving competitive temporal quality and target consistency. We attribute the moderate trade-off in frame quality to the substantial capacity gap between the $1.3$B Wan2.1 draft and the $33$B H3 backbone. Qualitative comparisons in Figure~\ref{fig:qualitative-summary} further confirm that Draft Forcing generalizes across distinct model representations and VAE spaces.

\subsection{Ablation Studies}
\label{sec:ablations}

To analyze the influence of each component, we conduct ablation studies on the 14B Self-Forcing target over the 128-prompt MovieGenVideoBench subset~\cite{polyak2024moviegen} under the CausVid protocol~\cite{yin2025causvid}. Unless otherwise specified, we use $\lambda_{\Delta}{=}0.5$, $\lambda_{\mathrm{LPIPS}}{=}2.0$, $\tau{=}0.25$, and $\mathcal{B}{=}1.6$.

\begin{table}[H]
\centering
\small
\setlength{\aboverulesep}{0pt}
\setlength{\belowrulesep}{0pt}
\setlength{\extrarowheight}{2.4pt}
\setlength{\tabcolsep}{3.5pt}
\begin{tabularx}{\textwidth}{ccCCCCCCC}
\toprule
$\lambda_{\Delta}$ & $\lambda_{\mathrm{LPIPS}}$ & \begin{tabular}[t]{@{}c@{}}Temporal\\Quality$\uparrow$\end{tabular} & DD$\uparrow$ & \begin{tabular}[t]{@{}c@{}}Frame\\Quality$\uparrow$\end{tabular} & \begin{tabular}[t]{@{}c@{}}Text\\Alignment$\uparrow$\end{tabular} & SSIM$\uparrow$ & PSNR$\uparrow$ & LPIPS$\downarrow$ \\
\midrule
0.5 & 0 & 90.33 & 53.91 & 65.27 & \textbf{28.61} & \textbf{0.6287} & \textbf{19.41} & 0.3175 \\
0 & 2 & 90.29 & 51.56 & 65.56 & 28.53 & 0.6189 & 19.01 & 0.3183 \\
\highlightrow
0.5 & 2 & \textbf{90.74} & \textbf{55.47} & \textbf{65.73} & 28.46 & 0.6200 & 19.14 & \textbf{0.3155} \\
\bottomrule
\end{tabularx}
\caption{Ablation study on the regularization weights ($\lambda_{\Delta}, \lambda_{\mathrm{LPIPS}}$) in the draft training objective. All variants are evaluated without the verifier by accepting one draft chunk after each target chunk; DD denotes Dynamic Degree.}
\label{tab:abl-loss}
\end{table}

\noindent\textbf{Draft Training Objective.} We study the influence of the loss terms in Eq.~\eqref{eq:draft-loss} without the verifier by accepting one draft chunk after each target chunk. As shown in Table~\ref{tab:abl-loss}, omitting $\mathcal{L}_{\mathrm{LPIPS}}$ ($\lambda_{\mathrm{LPIPS}}{=}0$) leads to a clear drop in visual sharpness, as reflected by the degradation in frame quality and LPIPS. Meanwhile, omitting $\mathcal{L}_{\Delta}$ ($\lambda_{\Delta}{=}0$) noticeably reduces DD, confirming the effectiveness of $\mathcal{L}_{\Delta}$ in supervising dynamic regions and temporal motion.

\begin{table}[H]
\centering
\small
\setlength{\aboverulesep}{0pt}
\setlength{\belowrulesep}{0pt}
\setlength{\extrarowheight}{2.4pt}
\setlength{\tabcolsep}{3.5pt}
\begin{tabularx}{\textwidth}{ccCCCCCCC}
\toprule
On-policy & Input & \begin{tabular}[t]{@{}c@{}}Draft\\Ratio\end{tabular} & \begin{tabular}[t]{@{}c@{}}Temporal\\Quality$\uparrow$\end{tabular} & \begin{tabular}[t]{@{}c@{}}Frame\\Quality$\uparrow$\end{tabular} & \begin{tabular}[t]{@{}c@{}}Text\\Alignment$\uparrow$\end{tabular} & SSIM$\uparrow$ & PSNR$\uparrow$ & LPIPS$\downarrow$ \\
\midrule
$-$ & $-$ & 42.86\% & 90.74 & \textbf{65.73} & 28.46 & 0.6200 & 19.14 & 0.3155 \\
\xmark & feature & 53.57\% & \textbf{91.07} & 65.47 & 28.48 & 0.6105 & 18.90 & 0.3299 \\
\cmark & latent & 52.90\% & 90.83 & 65.53 & 28.46 & 0.6701 & 21.70 & 0.2835 \\
\highlightrow
\cmark & feature & 51.12\% & 90.81 & 65.48 & \textbf{28.53} & \textbf{0.6818} & \textbf{22.44} & \textbf{0.2740} \\
\bottomrule
\end{tabularx}
\caption{Ablation study on the verifier training paradigm and input representation. ``$-$'' denotes accepting one draft chunk after each target chunk without the verifier. \cmark{} and \xmark{} denote on-policy rollout and off-policy trajectory supervision, respectively. All verifier-based variants are compared at a similar draft ratio.}
\label{tab:abl-scorer}
\end{table}

\noindent\textbf{Verifier Design and On-policy Training.} In Table~\ref{tab:abl-scorer}, our on-policy trained feature-space verifier raises the draft ratio over fixed periodic acceptance (first row) while substantially improving SSIM, PSNR, and LPIPS. Conversely, off-policy supervision on pre-recorded target trajectories yields no clear gain over the verifier-free baseline: once accepted draft chunks shift the history from the offline trajectory, the verifier biases toward lower scores on later chunks rather than actual chunk quality. Moreover, using the draft's final-layer hidden state (feature) outperforms the denoised latent (latent), confirming that transformer features offer richer verification cues.

\noindent\textbf{Inference Hyperparameters.} Figure~\ref{fig:abl-infer} illustrates the trade-off controlled by the acceptance threshold $\tau$ and the cumulative error budget $\mathcal{B}$. Lowering $\tau$ from $0.40$ to $0.10$ or relaxing $\mathcal{B}$ increases the draft ratio from $33.71\%$ to $67.19\%$. While accepting more draft chunks lowers SSIM, VBench Mean decreases only slightly.

%% file: sec/tables_lightx2v_main.tex
\begin{table}[t]
\centering
\footnotesize
\setlength{\tabcolsep}{1.6pt}
\renewcommand{\arraystretch}{1.12}
\begin{tabularx}{\textwidth}{l*{7}{c}CCC}
\toprule
& \multicolumn{4}{c}{VBench Score$\uparrow$} & \multicolumn{3}{c}{Visual Quality} & \multicolumn{3}{c}{Inference Efficiency} \\
\cmidrule(lr){2-5}\cmidrule(lr){6-8}\cmidrule(lr){9-11}
Method & Total & Quality & Semantic & \begin{tabular}[t]{@{}c@{}}Dynamic\\Degree\end{tabular} & SSIM$\uparrow$ & PSNR$\uparrow$ & LPIPS$\downarrow$ & \begin{tabular}[t]{@{}c@{}}Draft\\Ratio\end{tabular} & \begin{tabular}[t]{@{}c@{}}DiT\\Speedup$\uparrow$\end{tabular} & \begin{tabular}[t]{@{}c@{}}E2E\\Speedup$\uparrow$\end{tabular} \\
\midrule
Target-only (14B) & 84.55 & 85.18 & 81.99 & 75.83 & -- & -- & -- & -- & $1.00\times$ & $1.00\times$ \\
Training-free Draft & 84.57 & 85.21 & 82.02 & 74.44 & 0.5982 & 18.61 & 0.3268 & 42.86\% & $1.38\times$ & $1.35\times$ \\
SDVG~\cite{sdvg2026} & 82.76 & 82.93 & \textbf{82.08} & 65.83 & 0.4660 & 15.73 & 0.5273 & 74.31\% & $1.75\times$ & $1.67\times$ \\
\highlightrow
Draft Forcing & 84.41 & 85.12 & 81.53 & 75.00 & \textbf{0.6764} & \textbf{20.26} & \textbf{0.2635} & 56.77\% & $1.63\times$ & $1.57\times$ \\
\highlightrow
Draft Forcing$^{\dagger}$ & 84.35 & 85.07 & 81.49 & 75.28 & 0.6187 & 19.09 & 0.3199 & 70.79\% & $\mathbf{2.02\times}$ & $\mathbf{1.89\times}$ \\
\midrule
Forcing-KV~\cite{ji2026forcingkv} & \textbf{84.60} & 85.33 & 81.70 & \textbf{78.61} & 0.6469 & 20.10 & 0.2869 & -- & $1.18\times$ & $1.16\times$ \\
\highlightrow
\quad + Draft Forcing & 84.56 & \textbf{85.34} & 81.47 & 76.94 & 0.6154 & 19.03 & 0.3182 & 57.03\% & $1.84\times$ & $1.74\times$ \\
\midrule
Dummy Forcing~\cite{guo2026dummyforcing} & 84.54 & 85.21 & 81.88 & 76.94 & 0.5952 & 16.98 & 0.3579 & -- & $1.24\times$ & $1.22\times$ \\
\highlightrow
\quad + Draft Forcing & 84.53 & 85.25 & 81.68 & 76.39 & 0.5728 & 16.38 & 0.3719 & 57.53\% & $1.93\times$ & $1.82\times$ \\
\bottomrule
\end{tabularx}
\caption{Quality, fidelity, and efficiency on Krea Realtime 14B with a 1.3B draft on VBench. Fidelity metrics and speedups are measured relative to target-only. Rows shaded in green indicate Draft Forcing variants ($\tau{=}0.25$, $\mathcal{B}{=}1.6$ by default; $\tau{=}0.10$, $\mathcal{B}{=}3.0$ for $\dagger$).}
\label{tab:vbench-full}
\end{table}

%% file: sec/2_related.tex
\FloatBarrier
\section{Related Work}
\label{sec:related}

\noindent\textbf{Streaming Autoregressive Video Generation.} Streaming video diffusion models replace bidirectional denoising~\cite{ho2022videodiffusion,blattmann2023videoldm,wan2025} with chunk-wise causal generation and KV-cache reuse~\cite{chen2024diffusionforcing}. CausVid~\cite{yin2025causvid} distills bidirectional models into causal generators, while Self-Forcing~\cite{huang2025selfforcing} and follow-ups~\cite{causalforcing2026,causalforcingpp2026,oneforcing2026,causalrcm2026,krea2025realtime} bridge the train--test gap and compress per-chunk denoising to $1$--$4$ steps~\cite{salimans2022progressivedistillation,song2023consistency}. Minute-scale methods~\cite{henschel2025streamingt2v,liu2025rollingforcing,cui2025selfforcingpp,yang2025longlive} further extend generation length. However, they only accelerate the large model internally, which is still invoked sequentially per chunk.

\begin{wrapfigure}{r}{0.52\textwidth}
\centering
\includegraphics[width=\linewidth]{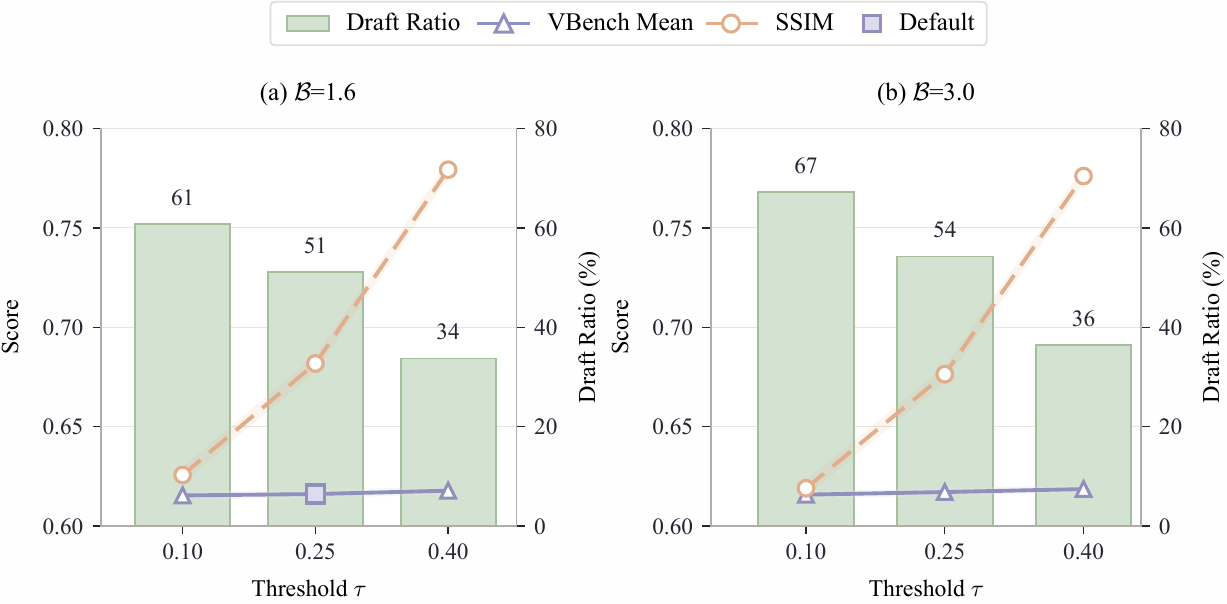}
\caption{Ablation study on the acceptance threshold $\tau\in\{0.10,0.25,0.40\}$ under cumulative error budgets (a)~$\mathcal{B}{=}1.6$ and (b)~$\mathcal{B}{=}3.0$. Bars show the draft ratio (right axis), while curves show SSIM and VBench mean (left axis), where VBench mean averages temporal quality, frame quality, and text alignment normalized to $[0,1]$. The square marker indicates the default setting ($\tau{=}0.25$, $\mathcal{B}{=}1.6$).}
\label{fig:abl-infer}
\end{wrapfigure}
\noindent\textbf{Speculative Decoding in Language Models.} Speculative decoding accelerates language generation by proposing candidate tokens with a lightweight draft and verifying them in one parallel target pass via rejection sampling~\cite{leviathan2023speculative,chen2023speculative}. Follow-up works improve drafting via multi-head decoding~\cite{cai2024medusa,gloeckle2024mtp}, Jacobi lookahead~\cite{fu2024lookahead}, on-policy distillation~\cite{zhou2024distillspec}, or adaptive lengths~\cite{huang2024specdecpp}. Feature-level drafters like EAGLE~\cite{li2024eagle,li2025eagle3}, DFlash~\cite{chen2026dflash}, and DFlare~\cite{dflare2026} condition drafts on multi-layer target hidden states with layer-wise fusion and heterogeneous KV projections. Yet, all rely on parallel target verification over discrete tokens, which is prohibitive for continuous multi-step video chunks.

\noindent\textbf{Speculative Acceleration for Video Generation.} Speculative video acceleration operates at the timestep or chunk level. Timestep-level methods~\cite{dts2026,flowcast2026,asdsv2025} draft coarse intra-chunk steps, yielding limited gains on few-step streaming models. At the chunk level, Next Forcing~\cite{nextforcing2026} and Next-Frame Diffusion~\cite{nfd2025} omit verification, while X-Cache~\cite{xcache2026} and Interactive Speculative Sampling~\cite{huang2026speculativeinteractive} reuse states within a single model~\cite{peebles2023dit} instead of draft--target routing. SDVG~\cite{sdvg2026} pairs a training-free draft with post-decode ImageReward~\cite{xu2023imagereward}, causing draft--target mismatch, VAE overhead, and trajectory drift. Draft Forcing resolves these via target-guided drafting on persisted target features and an on-policy feature-space verifier. Concurrently, Speculative Forcing~\cite{speculativeforcing2026} skips token-level steps inside the target model via feature-based convergence prediction, complementing our chunk-level routing.

%% file: sec/5_conclusion.tex
\section{Conclusion}
\label{sec:conclusion}

We introduce \textbf{Draft Forcing}, a chunk-level speculative framework coupling target-guided drafting with a learned verifier for streaming autoregressive video generation. To align the draft for high acceptance, it adopts feature-level drafting, conditioning on multi-layer target hidden states persisted from committed chunks to match target trajectories with multi-aspect losses. For verification, an on-policy feature-space verifier predicts CDF-calibrated frame-wise relative RMSE scores against the target. Trained along speculative rollouts as contexts anneal from random to inference-aligned mixtures, it gates proposals directly without VAE decoding. On 14B Self-Forcing, Draft Forcing achieves $1.63\times$ DiT and $1.57\times$ E2E speedups on VBench with minimal quality loss, reaching up to $1.93\times$ DiT ($1.82\times$ E2E) with orthogonal KV-cache compression. Moreover, it generalizes across model families to accelerate RAVEN-H3, generating over $50\%$ of frames with the draft while maintaining competitive quality.

%% file: sec/appendix.tex
\section{Speculative Inference Procedure}
\label{app:inference}

Algorithm~\ref{alg:draftforcing} details the complete speculative inference workflow of Draft Forcing. The generation starts with the target model $T$ producing and committing the opening chunk $x_1$. For each subsequent chunk $t \ge 2$ selected for drafting, the draft model $D$ proposes a candidate $\hat{x}_t$. The feature-space verifier $\mathcal{V}$ evaluates this candidate without VAE decoding. Proposals meeting the threshold $\tau$ are accepted directly, while rejected chunks trigger fallback regeneration by $T$. When consecutive draft acceptances cause the cumulative error budget $E_{\mathrm{cum}}$ to exceed $\mathcal{B}$, a target refresh is scheduled to prevent long-horizon drift before resuming speculation.

\begin{algorithm}[h]
\caption{Speculative Inference for Draft Forcing}
\label{alg:draftforcing}
\begin{algorithmic}[1]
\REQUIRE Target $T$; draft $D$; verifier $\mathcal{V}$; prompt $c$; chunk count $N$; threshold $\tau$; error budget $\mathcal{B}$.
\algstage{Initial chunk generation}
\STATE Generate $x_1$ with $T$. Set $\mathcal{H}_2=\{x_1\}$, $t_{\mathrm{anc}}=1$, and $E_{\mathrm{cum}}=0$.
\STATE Cache KV and hidden states $\{h^{(i)}_T(\mathcal{H}_2)\}_{i\in\mathcal{I}}$ with $T$.
\FOR{$t=2,\ldots,N$}
\IF{$E_{\mathrm{cum}} \ge \mathcal{B}$}
\STATE Generate $x_t$ with $T$ using $\mathcal{H}_t$. Set $t_{\mathrm{anc}}\leftarrow t$ and $E_{\mathrm{cum}}\leftarrow 0$.
\ELSE
\algstage{Target-guided drafting}
\STATE Propose $\hat{x}_t$ and $h_D(\hat{x}_t)$ with $D$ using $\{h^{(i)}_T(\mathcal{H}_t)\}_{i\in\mathcal{I}}$ (Eqs.~\ref{eq:fusion} and \ref{eq:het-kv}).
\algstage{Feature-space verification}
\STATE Predict frame scores $\{\hat{s}_{t,f}\}_{f=1}^{F}$ with $\mathcal{V}$ using $h_D(\hat{x}_t)$ and cached target features from $x_{t-1}$ and $x_{t_{\mathrm{anc}}}$.
\STATE Compute chunk score $\hat{s}_t=\min_{f}\hat{s}_{t,f}$ (Eq.~\ref{eq:chunk-score}).
\IF{$\hat{s}_t \ge \tau$}
\STATE Set $x_t \leftarrow \hat{x}_t$ and $E_{\mathrm{cum}} \leftarrow E_{\mathrm{cum}} + \sum_{f=1}^{F}\Phi_{\mathrm{train}}^{-1}(1-\hat{s}_{t,f})$.
\ELSE
\STATE Regenerate $x_t$ with $T$ using $\mathcal{H}_t$. Set $t_{\mathrm{anc}}\leftarrow t$ and $E_{\mathrm{cum}}\leftarrow 0$.
\ENDIF
\ENDIF
\algstage{History and cache update}
\STATE Update KV cache, $\{h^{(i)}_T(\mathcal{H}_{t+1})\}_{i\in\mathcal{I}}$, and $\mathcal{H}_{t+1}=\mathcal{H}_t\cup\{x_t\}$ with $T$.
\ENDFOR
\RETURN $\mathcal{H}_{N+1}$
\end{algorithmic}
\end{algorithm}

\section{Chunk-wise Routing and Acceptance Analysis}
\label{app:chunk-analysis}

\begin{wrapfigure}{r}{0.52\textwidth}
\centering
\includegraphics[width=\linewidth]{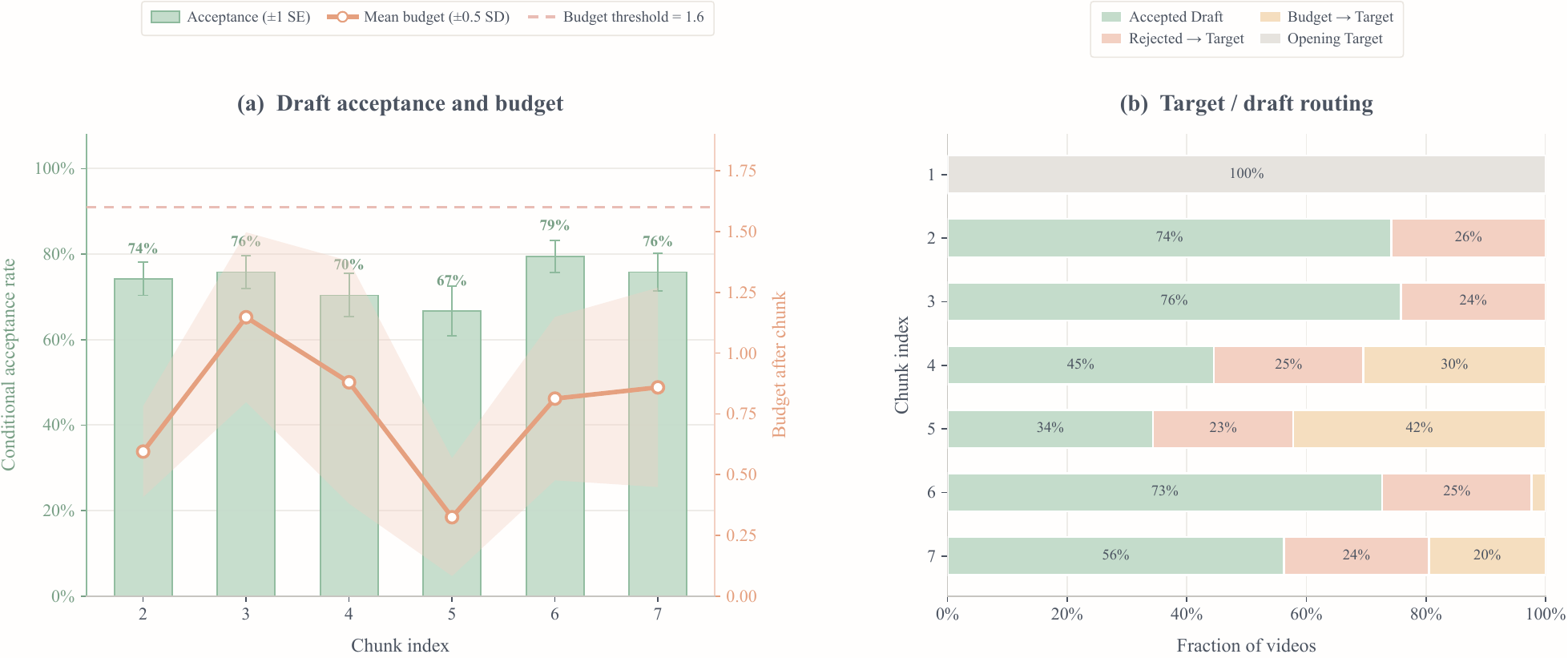}
\caption{Per-chunk routing and acceptance on the $128$-prompt MovieGenVideoBench subset at the default setting ($\tau{=}0.25$, $\mathcal{B}{=}1.6$). (a)~Verifier acceptance rate among scored draft proposals (distinct from the final video draft ratio; $\pm 1$ SE) and mean cumulative error budget ($\pm 0.5$ SD). (b)~Fraction of videos by producer and routing outcome at each chunk position.}
\label{fig:chunk-analysis}
\end{wrapfigure}
Figure~\ref{fig:chunk-analysis} analyzes per-chunk speculative routing and draft acceptance on the $128$-prompt MovieGenVideoBench subset~\cite{polyak2024moviegen} under the default setting ($\tau{=}0.25$, $\mathcal{B}{=}1.6$). Across chunk positions, the draft acceptance rate first decreases and then rises: it stays around $74.2\%$--$75.8\%$ at chunks $2$--$3$, dips to $66.7\%$ at chunk $5$, and rebounds to $79.5\%$ and $75.8\%$ at chunks $6$--$7$. This behavior is natural. Early in the sequence, accepting consecutive draft chunks accumulates approximation error, which raises the cumulative error budget $E_{\mathrm{cum}}$ and triggers target refreshes around the middle chunks (chunks $4$--$5$), lowering the acceptance rate. Later in the sequence, as more committed historical chunks accumulate in $\mathcal{H}_t$, the draft can better capture the motion trajectory and generate higher-quality proposals, causing the acceptance rate to rise again.

\section{Detailed Verifier Architecture and On-policy Training}
\label{app:verifier}

\subsection{Verifier Architecture}
To verify draft proposals with minimal overhead, the verifier $\mathcal{V}$ is designed as a lightweight FP32 cross-attention head ($8.27\text{M}$ parameters in the primary $14\text{B}$ Self-Forcing setting, with internal width $d_h{=}256$). Its computation is organized into three functional stages:
\begin{itemize}
\item \textbf{Compact feature encoding.} All chunk feature maps ($x_{t-1}$, $x_{t_{\mathrm{anc}}}$, and $h_D(\hat{x}_t)$) are first spatially pooled to a compact $5\times 8$ grid per frame. A shared target encoder normalizes, projects, and fuses the selected target layers $\mathcal{I}$ into width $d_h$, producing historical memories $M_{t-1}$ and $M_{\mathrm{anc}}$, while a matching projection maps $h_D(\hat{x}_t)$ to the draft query $Q_{\mathcal{V}}$.
\item \textbf{Dual-context discrepancy reasoning.} To assess both local transition continuity and multi-chunk drift, two parallel $8$-head cross-attention branches compare $Q_{\mathcal{V}}$ against $M_{t-1}$ and $M_{\mathrm{anc}}$. In the anchor branch, the chunk distance $g_t = t - t_{\mathrm{anc}} \ge 1$ is embedded from $\log(1+g_t)$ via a small MLP and added to $Q_{\mathcal{V}}$ so that $\mathcal{V}$ accounts for how far the rollout has progressed since the last target refresh.
\item \textbf{Frame-wise score readout.} The two attended contexts are concatenated, fused with $Q_{\mathcal{V}}$ through a residual MLP, spatially average-pooled per frame, and mapped through an output MLP with sigmoid activation to produce the calibrated score $\hat{s}_{t,f} \in (0, 1)$.
\end{itemize}

\subsection{On-policy Data Collection and CDF Calibration}
During on-policy verifier training, both $T$ and $D$ remain frozen, and only training prompts are sampled without any pre-recorded trajectories or offline labels:
\begin{enumerate}
\item \textbf{Rollout and paired target labeling.} Starting from the target-generated opening chunk $x_1$ (which initializes $\mathcal{H}_2$ and $x_{t_{\mathrm{anc}}}$), $D$ proposes $\hat{x}_t$ at each subsequent chunk $t \ge 2$ conditioned on $T$'s cached history, while its pooled final-layer hidden state $h_D(\hat{x}_t)$ is recorded. To obtain an exact same-history reference, $T$ generates a paired chunk $x^T_t$ from the identical committed history $\mathcal{H}_t$ and initial noise. This paired target pass is used solely to compute the frame-wise relative RMSE $e_{t,f}$ (Eq.~\eqref{eq:relative-rmse}) and does not modify $T$'s KV cache unless the proposal is rejected.
\item \textbf{Policy-driven history advancement.} During training, we commit the draft proposal $\hat{x}_t$ if its lowest predicted frame score meets the training threshold. Otherwise, we commit the paired target chunk $x^T_t$ and make it the new anchor. The committed chunk becomes part of the history for the next proposal. The cumulative error budget applies only at inference. Once reached, it triggers target generation on the next chunk.
\item \textbf{Continuous empirical CDF updates.} By default, $\mathcal{V}$ updates after every proposal so that subsequent chunks in the same rollout observe the updated verifier. Before each update, newly observed $e_{t,f}$ values are appended to the running pool of rollout RMSE samples and sorted to update $\Phi_{\mathrm{train}}$. Each $e_{t,f}$ is then mapped to its target score $s_{t,f} = 1 - \Phi_{\mathrm{train}}(e_{t,f})$ (Eq.~\eqref{eq:cdf-score}) via tie-aware mid-rank lookup, and $\mathcal{V}$ is updated with $\mathcal{L}_{\mathrm{Verifier}}$ (Eq.~\eqref{eq:scorer-loss}). Once training finishes, the final empirical CDF is frozen in the verifier checkpoint for inference.
\end{enumerate}

\section{Verifier Score Calibration and Error Prediction}
\label{app:rmse-analysis}

\begin{figure}[t]
\centering
\includegraphics[width=\textwidth]{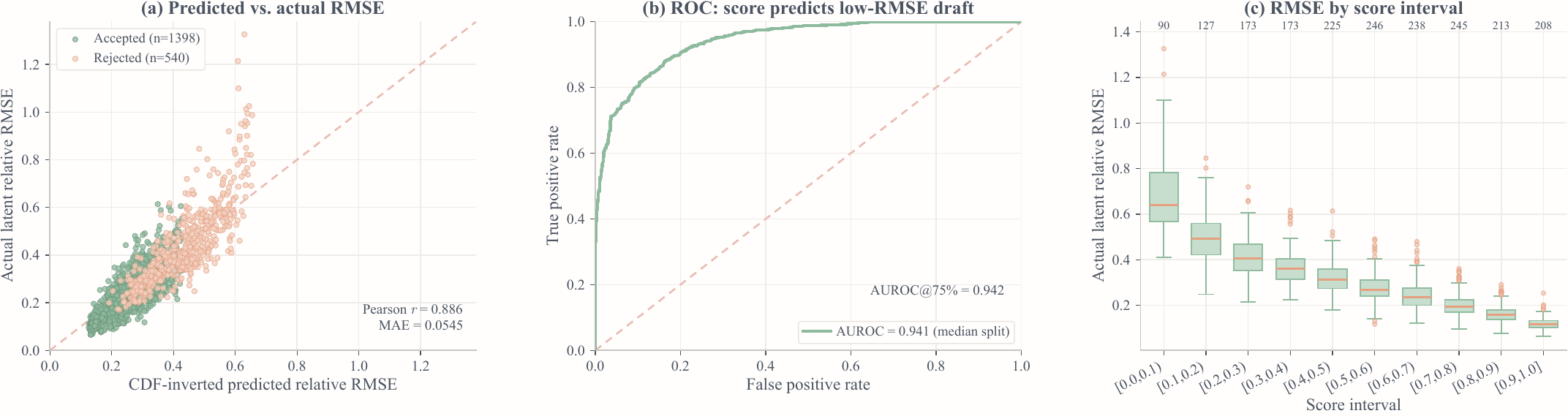}
\caption{Verifier error prediction across all $1{,}938$ scored draft frames (both accepted and rejected proposals) on the $128$-prompt MovieGenVideoBench subset. (a)~CDF-inverted predicted relative RMSE versus actual latent relative RMSE against the paired target. (b)~ROC curve for identifying below-median RMSE frames from the predicted score, with AUROC$@75\%$ also reported. (c)~Distribution of actual latent relative RMSE across predicted score intervals.}
\label{fig:rmse-analysis}
\end{figure}

Figure~\ref{fig:rmse-analysis} evaluates the verifier's latent error prediction across all $1{,}938$ scored draft frames on the $128$-prompt MovieGenVideoBench subset~\cite{polyak2024moviegen}. Overall, the verifier fits the actual latent relative RMSE well: inverting the predicted scores through the empirical CDF $\Phi_{\mathrm{train}}$ yields a Pearson correlation of $r{=}0.886$ ($\mathrm{MAE}{=}0.0545$), separates low-error frames cleanly ($\mathrm{AUROC}{=}0.941$ at the median RMSE), and orders actual errors monotonically across score intervals (Figures~\ref{fig:rmse-analysis}b and~\ref{fig:rmse-analysis}c). In Figure~\ref{fig:rmse-analysis}a, however, a small tail of frames with high actual RMSE is under-predicted by the verifier, leaving some degraded proposals close to the threshold. This motivates accumulating the estimated frame errors $E_{\mathrm{cum}}$ across consecutive accepted draft chunks so that target regeneration is triggered before residual errors compound over long sequences.

\section{Training and Architecture Details}
\label{app:training}

\subsection{Draft and Verifier Training} We evaluate Draft Forcing on two streaming autoregressive video diffusion backbones: (1) our primary same-backbone setting using the 14B Self-Forcing model~\cite{huang2025selfforcing} (Krea Realtime 14B~\cite{krea2025realtime}) as the frozen target $T$, and (2) a cross-backbone setting using RAVEN-H3, the RAVEN streaming adaptation~\cite{lu2026raven} of MiniMax-H3~\cite{minimax2026h3}, as the frozen target $T$. To reduce training cost and inherit pretrained motion and appearance priors, in both settings we initialize the draft $D$ from the pretrained 1.3B Self-Forcing model~\cite{huang2025selfforcing} ($30$ layers). In the primary Self-Forcing setting, we collect $10\text{k}$ clean target trajectories from the VidProM prompt training split~\cite{wang2024vidprom}, where each video consists of $N{=}7$ chunks of $F{=}3$ latent frames ($21$ latent frames in total, decoded to $81$ RGB frames at $480\times 832$ resolution and $16\text{ fps}$).

The draft uses four logical timesteps $(1000, 750, 500, 250)$. The shift in Appendix~\ref{app:flow-schedule} ($\gamma=5$) maps them to approximately $(1000, 937.5, 833.33, 625)$. Draft training uses the shifted values, and our inference implementation truncates them to the integer model timesteps $(1000, 937, 833, 625)$. Krea Realtime 14B generates each target chunk with five denoising passes at model timesteps $(1000, 937, 833, 625, 0)$. For each nonfinal chunk, the target performs a separate forward pass at timestep $0$ on the selected clean latent to update its KV cache.

We tune $D$'s self-attention and dedicated context KV projections with LoRA~\cite{hu2021lora} using a rank of $32$ and a scaling factor of $32$. For the training objective in Eq.~\eqref{eq:draft-loss}, we decode $20\times 20$ random spatial latent crops through the frozen Wan~\cite{wan2025} VAE for $\mathcal{L}_{\mathrm{LPIPS}}$~\cite{zhang2018lpips}, set $\lambda_{\Delta}{=}0.5$ and $\lambda_{\mathrm{LPIPS}}{=}2.0$, and train the draft for $22.5\text{k}$ steps using an AdamW optimizer~\cite{loshchilov2019adamw} with a learning rate of $2\times 10^{-4}$.

We then freeze both models and train the verifier for $700$ on-policy rollout iterations using AdamW with a learning rate of $10^{-4}$. During training, a draft chunk is accepted if its lowest frame score is at least $0.50$. At inference, our default setting uses an acceptance threshold of $0.25$ and a cumulative error budget of $1.6$. Architecturally, $\mathcal{V}$ extracts target historical features from $\mathcal{I}{=}\{4,12,20,28,36\}$.

\subsection{Cross-backbone Adaptation}

\begin{figure}[H]
\centering
\includegraphics[width=\textwidth]{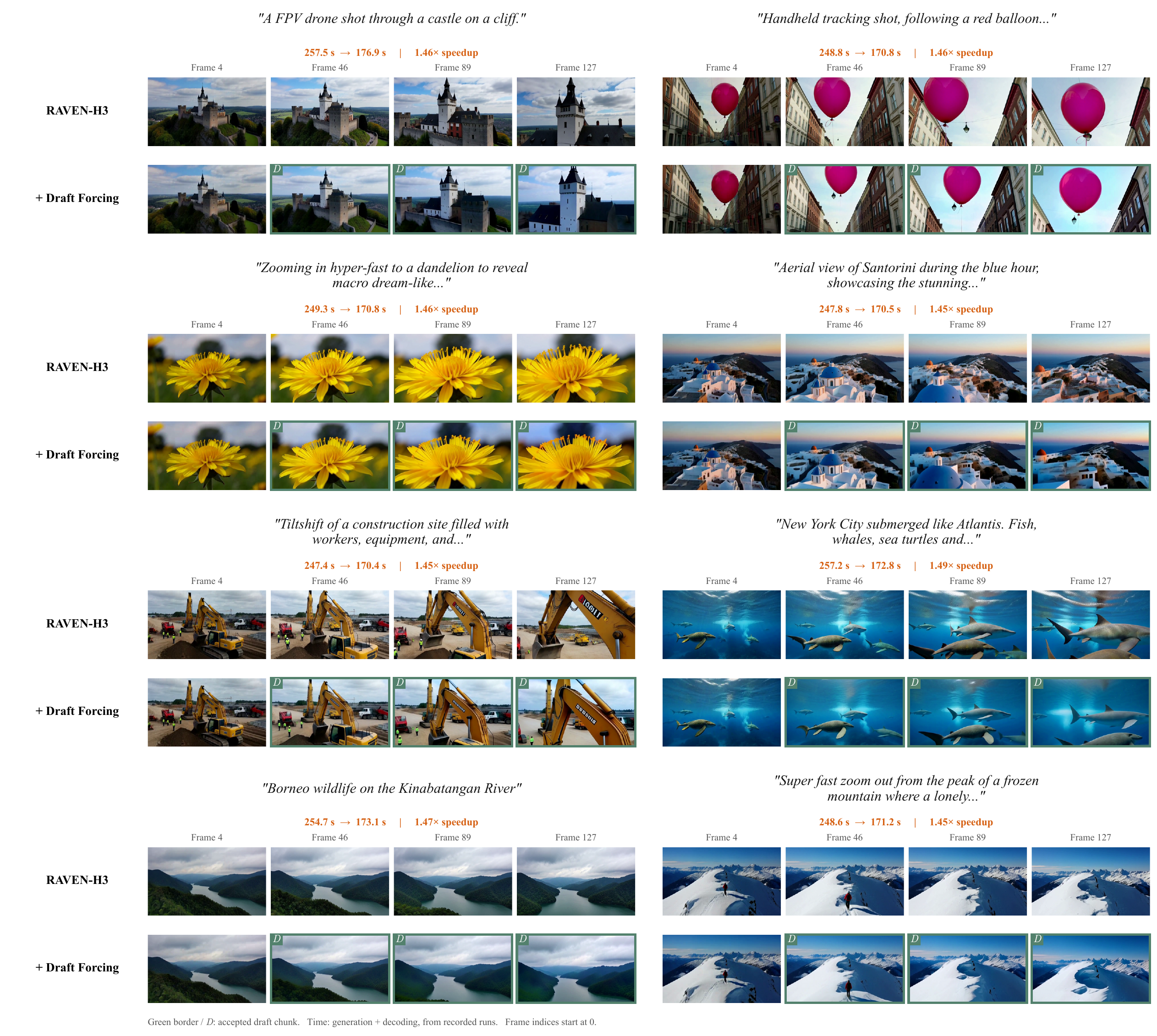}
\caption{Qualitative results of accelerating RAVEN-H3 with Draft Forcing.}
\label{fig:h3-qual}
\end{figure}

In the cross-backbone RAVEN-H3 setting (Section~\ref{sec:cross-model}), each video comprises $N{=}8$ full chunks of $F{=}5$ latent frames plus a $2$-frame tail chunk ($141$ RGB frames at $480\times 832$ resolution and $24\text{ fps}$, ${\sim}5.9\text{ s}$). Both the Target-only RAVEN-H3 baseline and our accelerated pipeline skip audio synthesis and operate in video-only mode, because no smaller streaming joint audio--video checkpoint is available to initialize the draft model. To translate Wan draft latents ($16\times 5\times 60\times 104$) into H3 target latents ($24\times 5\times 30\times 52$), the latent mapper $\mathcal{M}$~\cite{li2026solh3} packs $2\times 2$ spatial neighborhoods via $\mathrm{PixelUnshuffle}$ into channels and passes them through a ridge-initialized frozen $1\times 1\times 1$ affine skip and $22$ factorized 3D residual blocks of width $752$. We pretrain $\mathcal{M}$ for $20\text{k}$ steps using latent $L_2$ loss together with $L_1$ ($0.25$), temporal ($0.10$), and spatial ($0.02$) finite-difference losses, followed by $7.5\text{k}$ steps of decoder-aware supervision with $\lambda_{\mathrm{dec}}{=}160$ in Eq.~\eqref{eq:mapper-loss}. Freezing $\mathcal{M}$, we then train the Wan draft for $8.5\text{k}$ steps (with the same optimizer and loss weights, using a $2\text{k}$-step warmup for $\mathcal{L}_{\mathrm{LPIPS}}$). We then train the RAVEN-H3 verifier for $540$ on-policy rollout iterations with an acceptance threshold of $0.25$. The default inference policy retains this threshold and adds a cumulative error budget of $1.6$. Figure~\ref{fig:h3-qual} shows qualitative results of accelerating RAVEN-H3 with Draft Forcing.

\subsection{Evaluation and Timing Protocol}
\label{app:measurement}

We follow the standard VBench protocol to evaluate generation quality. Efficiency is measured on NVIDIA H200 GPUs with batch size $1$ using matched prompts and seeds. DiT time covers drafting, verification, fallback target generation, and KV operations; E2E time spans text encoding, DiT sampling, synchronous VAE decoding, and RGB transfer to CPU.

\subsection{Detailed Feature Fusion Module}
\label{app:fuse}

In each draft layer $\ell$, $\mathrm{Fuse}_\ell$ (Eq.~\eqref{eq:fusion}) computes a softmax-weighted mixture over cached target hidden states from layers $\mathcal{I}{=}\{4,12,20,28,36\}$, followed by a linear width projection and RMSNorm. Because the draft backbone is relatively deep, shallow, intermediate, and deep blocks operate at different abstraction levels and require distinct mixtures of target features; sharing a single fused context across all draft layers is therefore suboptimal. Instead, for an $L_D$-layer draft and $|\mathcal{I}|$ selected target layers, each draft layer $\ell$ learns its own $|\mathcal{I}|$ mixing weights (giving $L_D \times |\mathcal{I}|$ mixing weights in total) so that every block can emphasize the target depths most informative for its own representations:
\begin{equation}
f^{(\ell)}_{\mathrm{ctx}} = \mathrm{RMSNorm}\!\left(\left(\sum_{i \in \mathcal{I}} \alpha^{(\ell)}_i \, h^{(i)}_T(\mathcal{H}_t)\right) W^{(\ell)}_{\mathrm{proj}}\right),
\end{equation}
where $\alpha^{(\ell)} = \mathrm{softmax}(w^{(\ell)})$. In this base module, the learnable parameters per draft layer are the zero-initialized mixing logits $w^{(\ell)} \in \mathbb{R}^{|\mathcal{I}|}$ and the linear width projection $W^{(\ell)}_{\mathrm{proj}} \in \mathbb{R}^{d_T \times d_D}$ (mapping from target width $d_T$ to draft width $d_D$), while $\mathrm{RMSNorm}$ is parameter-free. The fused context $f^{(\ell)}_{\mathrm{ctx}}$ is then fed into the dedicated context KV projections in Eq.~\eqref{eq:het-kv}, with causal RoPE applied to context keys from frame $0$ and to draft queries and keys from the current chunk's frame offset before concatenating them with the current-chunk KV for attention.

To align H3 historical features ($5376$ channels, $15\times 26$ spatial grid) from layers $\mathcal{I}{=}\{8,20,32,44\}$ with the Wan 1.3B draft ($1536$ channels, $30\times 52$ grid), we augment $\mathrm{Fuse}_\ell$ with a shared spatiotemporal adapter ($384$ bottleneck channels), a $2\times$ $\mathrm{PixelShuffle}$~\cite{shi2016subpixel} spatial alignment branch, $128$-dimensional token-adaptive layer gating, and block-specific residual MLPs. Cross-backbone $\mathrm{Fuse}_\ell$ therefore retains the exact same per-layer mixing, width projection, and $\mathrm{RMSNorm}$ backbone above.
\begin{itemize}
\item \textbf{Pre-fusion spatial alignment.} Before per-layer mixing, a shared spatiotemporal adapter normalizes the native target features, refines them in a compact bottleneck with depthwise spatial and causal temporal convolutions, and upsamples them to the draft's spatial grid via trilinear interpolation with a $\mathrm{PixelShuffle}$ residual branch.
\item \textbf{Token-adaptive gating and refinement.} Inside each layer $\ell$, the static logits $w^{(\ell)}$ are modulated per token via dot-product gating with learnable gating features extracted from the aligned target states, and a zero-initialized residual MLP is appended after $\mathrm{RMSNorm}$ to refine the projected context.
\end{itemize}

\section{Noise Schedule and Flow-Matching Parameterization}
\label{app:flow-schedule}

Following Wan2.1~\cite{wan2025} and Self-Forcing~\cite{huang2025selfforcing}, we adopt a shifted flow-matching schedule for draft training and inference. For each training chunk $x^T_t$, we uniformly sample a logical denoising step $k \in \{1000, 750, 500, 250\}$ and compute the shifted noise level with shift factor $\gamma{=}5$:
\begin{equation}
\sigma_k = \frac{\gamma (k / 1000)}{1 + (\gamma - 1)(k / 1000)},
\end{equation}
feeding the warped timestep $1000\,\sigma_k$ to the draft model. Given Gaussian noise $\epsilon \sim \mathcal{N}(0, I)$, the noisy latent chunk is constructed by linear interpolation:
\begin{equation}
x^T_{t,k} = (1 - \sigma_k)x^T_t + \sigma_k\epsilon,
\end{equation}
with the ground-truth velocity target $\epsilon - x^T_t$ used in Eq.~\eqref{eq:flow-loss}.

\section{Full VBench Results}
\label{app:full-vbench}

Table~\ref{tab:vbench-all-dimensions} reports the full $16$-dimension VBench breakdown for Krea Realtime 14B, expanding on Table~\ref{tab:vbench-full}. Under the default setting, Draft Forcing closely matches the target-only baseline across the quality dimensions ($85.12$ vs.\ $85.18$), even slightly improving temporal flickering and motion smoothness. The largest drops occur on color ($-1.24$) and multiple objects ($-0.87$).

\begin{table}[H]
\centering
\small
\setlength{\tabcolsep}{2pt}
\renewcommand{\arraystretch}{1.10}
\begin{tabular*}{\textwidth}{@{\extracolsep{\fill}}l*{9}{c}@{}}
\toprule
VBench dimension & Target-only & \shortstack{Training free\\draft} & SDVG & DF & DF$^{\dagger}$ & Forcing-KV & \shortstack{Forcing-KV\\+ DF} & Dummy & \shortstack{Dummy\\+ DF} \\
\midrule
Subject consistency & 95.79 & 96.09 & 91.45 & 95.58 & 95.33 & 95.88 & 95.79 & 95.54 & 95.55 \\
Background consistency & 95.27 & 95.70 & 92.91 & 94.92 & 94.72 & 95.12 & 94.95 & 94.88 & 94.86 \\
Temporal flickering & 98.34 & 98.53 & 98.65 & 98.65 & 98.72 & 98.40 & 98.72 & 98.84 & 98.86 \\
Motion smoothness & 98.21 & 98.16 & 98.15 & 98.40 & 98.59 & 98.26 & 98.45 & 98.28 & 98.42 \\
Dynamic degree & 75.83 & 74.44 & 65.83 & 75.00 & 75.28 & 78.61 & 76.94 & 76.94 & 76.39 \\
Aesthetic quality & 66.61 & 66.43 & 65.57 & 66.24 & 65.93 & 66.33 & 66.28 & 66.28 & 66.13 \\
Imaging quality & 70.25 & 70.03 & 69.28 & 69.90 & 69.40 & 69.83 & 69.66 & 69.45 & 69.58 \\
Object class & 96.74 & 96.69 & 96.22 & 96.50 & 96.34 & 96.19 & 96.52 & 96.41 & 96.30 \\
Multiple objects & 91.46 & 91.54 & 87.45 & 90.59 & 90.29 & 91.59 & 90.44 & 90.98 & 90.67 \\
Human action & 97.80 & 98.20 & 99.00 & 97.60 & 97.80 & 97.60 & 97.80 & 98.20 & 98.20 \\
Color & 82.38 & 82.31 & 87.21 & 81.14 & 81.99 & 82.07 & 81.93 & 82.10 & 81.90 \\
Spatial relationship & 86.16 & 86.11 & 84.97 & 85.79 & 85.32 & 85.47 & 85.46 & 86.45 & 85.47 \\
Scene & 57.49 & 57.37 & 57.65 & 56.76 & 56.57 & 56.95 & 55.92 & 57.17 & 57.14 \\
Temporal style & 24.77 & 24.75 & 24.66 & 24.67 & 24.65 & 24.71 & 24.70 & 24.75 & 24.73 \\
Appearance style & 20.27 & 20.32 & 20.37 & 20.32 & 20.33 & 20.20 & 20.26 & 20.14 & 20.18 \\
Overall consistency & 27.13 & 27.13 & 27.20 & 27.03 & 26.94 & 27.16 & 27.06 & 27.22 & 27.13 \\
\midrule
Total & 84.55 & 84.57 & 82.76 & 84.41 & 84.35 & 84.60 & 84.56 & 84.54 & 84.53 \\
Quality & 85.18 & 85.21 & 82.93 & 85.12 & 85.07 & 85.33 & 85.34 & 85.21 & 85.25 \\
Semantic & 81.99 & 82.02 & 82.08 & 81.53 & 81.49 & 81.70 & 81.47 & 81.88 & 81.68 \\
\bottomrule
\end{tabular*}
\caption{VBench scores for all $16$ dimensions and their aggregates ($0$--$100$). Column headers abbreviate Draft Forcing (DF) and Dummy Forcing (Dummy). DF uses $\tau{=}0.25$, $\mathcal{B}{=}1.6$ by default; DF$^{\dagger}$ uses $\tau{=}0.10$, $\mathcal{B}{=}3.0$ with full KV caches. Each configuration generates $6{,}220$ videos, including temporal-flickering filter candidates; each dimension follows its corresponding prompt subset and filtering rule.}
\label{tab:vbench-all-dimensions}
\end{table}

\Needspace{0.72\textheight}
\section{Accelerating RAVEN-H3}
\label{app:h3-qual}

